\PassOptionsToPackage{table}{xcolor}
\documentclass[10pt,twocolumn,letterpaper]{article}

\usepackage{iccv}      % To produce the REVIEW version
\usepackage{graphicx}
\usepackage{amsmath}
\usepackage{amssymb}
\usepackage{booktabs}
\usepackage{multirow}
\usepackage{array}
\usepackage{placeins}
\usepackage{algorithm}
\usepackage{algorithmic}

\definecolor{iccvblue}{rgb}{0.21,0.49,0.74}
\usepackage[pagebackref,breaklinks,colorlinks,allcolors=iccvblue]{hyperref}

\def\paperID{*****} % *** Enter the Paper ID here
\def\confName{ICCV}
\def\confYear{2025}

\title{From Scene-Centric to Observer-Centric: \\Modeling Observer-Aware Relations for 3D Scene Graph Generation}

\author{
    Jingjun Sun\textsuperscript{1}\thanks{Equal contribution.} \quad
    Chaowei Wang\textsuperscript{1}\footnotemark[1] \quad
    Zhirui Liu\textsuperscript{2}\footnotemark[1] \quad
    Jiaxu Tian\textsuperscript{1} \\ % 在这里换行，保持排版美观
    Ming Yang\textsuperscript{1} \quad
    Yaoxing Wang\textsuperscript{1} \quad
    Yan Di\textsuperscript{3} \quad
    Shan Gao\textsuperscript{1}\thanks{Corresponding author.} \\
    \textsuperscript{1}Northwestern Polytechnical University, Xi'an, Shaanxi, China\\
    \textsuperscript{2}ShanghaiTech University, Shanghai, China\\
    \textsuperscript{3}Harbin Institute of Technology, Harbin, Heilongjiang, China\\
    {\tt\small \{sunjingjun, chaowei\_wang, jiaxutian, yangziming, wangyx24\}@mail.nwpu.edu.cn},\\
    {\tt\small liuzhr2025@shanghaitech.edu.cn} ,
    {\tt\small diyan@hit.edu.cn} ,
    {\tt\small gaoshan@nwpu.edu.cn}
}

\begin{document}

% Teaser Figure (Placed before the abstract)
\twocolumn[{%
\renewcommand\twocolumn[1][]{#1}%
\maketitle
\vspace{-2.5em} % Adjust this spacing if it overlaps with the title
\begin{center}
    \includegraphics[width=\textwidth]{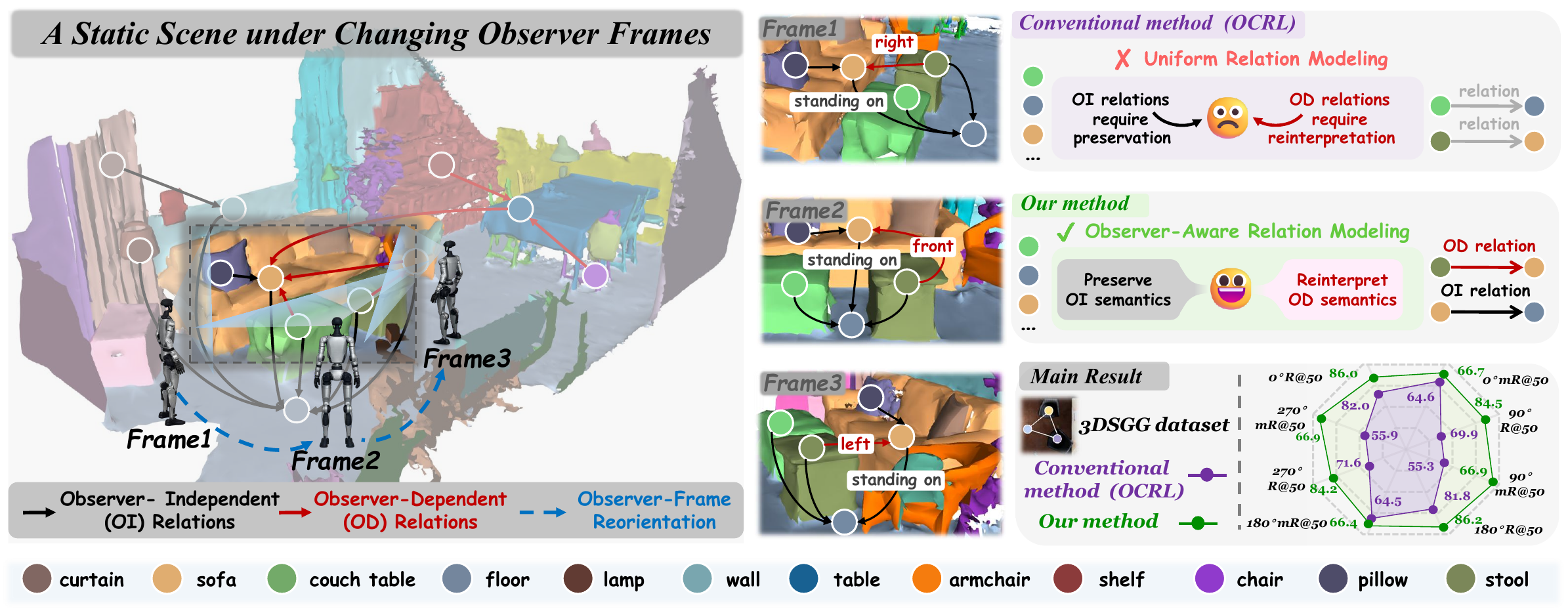}
    \captionof{figure}{Observer-Independent Relations, such as \textit{standing on}, preserve their
meanings across frames, whereas Observer-Dependent Relations, such
as \textit{right}, \textit{front}, and \textit{left}, are reinterpreted
relative to the current frame.
Conventional relation models \cite{heo2025object} treat both types uniformly, yielding
inconsistent predictions under observer-frame reorientation.
Our {Observer-Aware Relations} (\texttt{OAR}) account for their distinct frame responses
and improve relation prediction across the evaluated frames.}
    \label{fig:intro_rotation_motivation}
\end{center}
\vspace{1em}
}]

\begin{abstract}
3D Scene Graph Generation (3DSGG) represents 3D scenes as structured object--relation--object graphs for spatial understanding. 
In observer-centric spatial perception, the same scene may be expressed under different local observer frames while its structure remains unchanged. However, existing models typically assume a fixed scene-aligned reference frame and may produce semantically inconsistent predictions when the scene is re-expressed in another observer frame.
We attribute this failure to the heterogeneous frame dependency of relational predicates. 
Directional predicates such as \textit{left}, \textit{front}, \textit{right}, and \textit{behind} are \textbf{Observer-Dependent Relations}, whereas most contact, support, and semantic predicates, such as \textit{standing on} and \textit{attached to}, are approximately \textbf{Observer-Independent Relations}. 
Conventional models do not distinguish these frame responses, leading to degraded relation prediction under observer-frame reorientation.
We introduce \textbf{Observer-Aware Relations (OAR)}, which combines observer-aware geometric encoding and relation specialization, supported by frame-stable object encoding, for unified multi-label predicate prediction. 
Experiments on 3DSSG show that OAR consistently outperforms baselines across controlled observer-frame reorientations without training-time frame-reorientation augmentation, while remaining competitive on the standard benchmark.
The project page is available at \url{https://oar-predicate.github.io/}.
\end{abstract}

\section{Introduction}

Understanding a 3D scene requires not only recognizing individual objects, but also capturing the relational structure among them \cite{Armeni20193DSG,wald2020learning}. 3D Scene Graph Generation (3DSGG) addresses this challenge by representing a scene as a graph: objects are modeled as nodes, and their spatial relationships are modeled as edges. 
By transforming raw 3D observations into graph-based scene representations, 3DSGG serves as a fundamental step toward spatial reasoning and embodied intelligence \cite{Rosinol20203DDS,gu2024conceptgraphs,kim2024openvla,liu2025commanding,bjorck2025gr00t}, providing high-level scene abstractions for downstream applications such as robotic navigation and augmented reality \cite{Rosinol20203DDS,Tahara2020RetargetableAC}.

Driven by the demands of these downstream applications, a rich body of 3DSGG frameworks has been developed. Most 3DSGG methods ~\cite{wald2020learning,wu2021scenegraphfusion,zhang2021exploiting,wang2023vl,gu2024conceptgraphs,koch2024open3dsg,heo2025object,zhang2021knowledge} infer object relations from pairwise object features and relative geometry, increasingly enhanced by visual-semantic priors and instance-level recognition. 
Evaluated under the conventional fixed-reference-frame setting, these methods have demonstrated strong performance in parsing complex 3D environments. 
However, this fixed-reference-frame assumption does not always hold in practical observer-centric applications, where the same physical scene may be perceived and represented under different local observer frames even though its underlying structure remains unchanged.
Such frame variations require relation prediction to account for how each predicate depends on the current observer frame.
For example, directional relations such as \textit{left} and \textit{front} vary with the observer frame, whereas structural relations such as \textit{attached to} do not.
Existing 3DSGG frameworks \cite{heo2025object,wang2023vl} largely overlook this issue and encode relations with different observer-frame dependencies within a shared representation space. As a result, relation cues governed by different frame dependencies become entangled in a unified representation, leading to substantial performance degradation under observer-frame reorientation.

% 这里不用说这么细，直接承接上一句，他们都忽略了这个问题，在变视角环境下，之前的方法结果会极具下降。(因为他们将所有谓词在同一特征空间学习)
% \textcolor{red}{To examine the consequence of this assumption, we evaluate existing models under controlled observer-frame reorientations. Our experiments reveal clear performance degradation under frame changes: while the models remain relatively stable under axis-preserving direction reversals, their performance drops substantially when the two horizontal axes are exchanged. A predicate-level analysis further shows that different relations respond differently to observer-frame reorientation.}

To better characterize these heterogeneous dependencies, we draw inspiration from cognitive neuroscience, which suggests that spatial cognition represents spatial information across multiple reference frames, including egocentric and allocentric ones~\cite{burgess2008spatial,committeri2004reference}. Following this broader view, we categorize 3D predicates by how their labels transform under observer-frame reorientation.

% As illustrated in Fig.~\ref{fig:intro_rotation_motivation}, relations such as \textit{standing on} and \textit{attached to} are \textbf{Observer-Independent Relations}: their meanings are grounded in the physical scene structure and should remain stable under observer-frame reorientation. In contrast, relations such as \textit{left}, \textit{right}, \textit{front}, and \textit{behind} are \textbf{Observer-Dependent Relations}: their meanings are defined relative to the current observer frame and should be reinterpreted when the frame changes.
As illustrated in Fig.~\ref{fig:intro_rotation_motivation}, we define relations grounded in objective physical structures, such as \textit{standing on} and \textit{attached to}, as \textbf{Observer-Independent Relations}. As allocentric relational predicates, their semantic labels are independent of the orientation of the observer frame and should therefore remain invariant under observer-frame reorientation. In contrast, directional relations such as \textit{left}, \textit{right}, \textit{front}, and \textit{behind} are formulated as egocentric relational predicates, which we define as \textbf{Observer-Dependent Relations}. Because their semantics are defined relative to the orientation of the observer's local frame, observer-frame reorientation induces a predictable transformation of their labels rather than mere preservation.
% 这一段缩一下，直接说之前的方法没有显式区分，会造成。。。。（直接不说也行。）
% \textcolor{red}{Conventional relation models do not explicitly account for these distinct transformation behaviors, instead encoding predicates with different observer-frame dependencies through a shared relation representation. Such an undifferentiated representation must simultaneously preserve observer-independent information and support frame-consistent transformations for observer-dependent relations, posing a representation challenge under observer-frame reorientation.}
% Based on this formulation, we propose \textbf{OAR} (\textbf{Observer-Aware Relations}), a 3DSGG framework that explicitly models predicate-specific observer-frame dependency. OAR consists of a frame-stable object encoder, observer-aware geometric encoding, and relation specialization. The object encoder extracts rotation-robust object features, while the geometric encoder constructs dependency-specific geometric cues. Relation specialization further employs non-shared relation GNNs with observer-dependency supervision to learn complementary representations. These representations are fused before a unified classifier predicts the full multi-label predicate space, avoiding a hard partition between relation categories.
Based on this formulation, we propose \textbf{OAR} (\textbf{Observer-Aware Relations}), a 3DSGG framework that explicitly models predicate-specific dependencies on the observer frame. OAR comprises a frame-stable object encoder, observer-aware geometric encoding, and relation specialization. The frame-stable object encoder extracts rotation-stable representations, while the geometric encoder constructs dependency-specific relation cues. Relation specialization further employs non-shared relation GNNs to capture distinct reasoning patterns for observer-independent and observer-dependent relations. Observer-dependency supervision encourages pathway representations to remain complementary. Finally, these representations are fused and passed to a unified classifier over the complete multi-label predicate vocabulary, explicitly modeling heterogeneous observer-frame dependencies without imposing a hard partition between relation categories.

Our core contributions are threefold:

\noindent\textbf{1)} We identify heterogeneous observer-frame dependency as an overlooked issue in 3D Scene Graph Generation, showing that some predicates are observer-independent and should remain stable, whereas others are observer-dependent and should be reinterpreted with the observer frame.

\noindent\textbf{2)}  We propose OAR, an observer-aware 3DSGG framework. Supported by frame-stable object encoding, OAR employs observer-aware geometric encoding and relation specialization to explicitly model predicate-specific observer-frame dependency while preserving unified multi-label prediction.

\noindent\textbf{3)} Extensive experiments under controlled observer-frame reorientations show that, without frame-reorientation augmentation, OAR outperforms the strongest augmented baseline by up to 8.5 Overall R@50 points under challenging cross-axis reorientations, while achieving the best SGCls and PredCls mean recall among the compared methods on the standard 3DSSG benchmark.

% Extensive experiments under controlled observer-frame reorientations demonstrate that OAR consistently outperforms strong baselines without relying on frame-reorientation augmentation, while remaining competitive on the standard 3DSGG benchmark.

%Extensive experiments under controlled observer-frame reorientations show that, without frame-reorientation augmentation, OAR outperforms the strongest augmented baseline by up to 8.5 Overall R@50 points under challenging cross-axis reorientations, while achieving the best SGCls and PredCls mean recall among the compared methods on the standard 3DSSG benchmark.

\section{Related Work}

\subsection{3D Scene Graph Generation}
% 修改意见（LZR）：
% 1.两段式没问题，但第一段太长，早期工作可以直接一句话带过，不要每个单独提了，在early approach后面直接一个括号引用最经典的三四篇就行了。
% 然后最近2年的相关工作可以提名字，但表述也要简洁，一句话概括每篇工作干了啥新的事情/新的方法范式解决这个问题。
% 第二段不要说别人做了啥，然后相反的我们做了啥。
% 而应该说：但他们都忽略了...的重要性，我们认识到了这个问题的重要性，并通过...解决了这个问题。然后第二段结束。整个长度应该缩减到现在的60-70%左右。一般rework每一段都是“头重脚轻”就是说第一段回顾相关工作占70%，然后第二段体现我们相比之前的牛逼的地方在哪占30%
3D scene graph generation extends 2D SGG~\cite{zellers2018neural,xu2017scene,tang2019learning} to 3D point clouds, aiming to construct structured object-relation-object graphs from 3D scene geometry.
Early 3DSGG methods established benchmark datasets and GNN-based prediction frameworks with instance-level point features and pairwise geometric descriptors~\cite{wald2020learning,wald2019rio,wu2021scenegraphfusion,zhang2021exploiting,zhang2021knowledge}.
Recent methods have advanced 3DSGG from different perspectives.
VL-SAT~\cite{wang2023vl} introduced visual-linguistic semantics assisted training, where a multi-modal oracle transfers visual and linguistic knowledge to a 3D model during training while inference uses only 3D inputs.
OCRL~\cite{heo2025object} emphasized discriminative object representation learning via object-centric contrastive pretraining, showing that stronger object features benefit both object classification and relationship prediction.
Other recent efforts improve RGB-based 3DSGG with segmentation-guided feature aggregation and statistical confidence rescoring~\cite{Yeo2025StatisticalCR}, or explore scene-graph-level alignment for downstream registration~\cite{DebSarkar2023SGAligner3S,Xie2024SGPGMPG,Chen2026OpenSGAE3}.
Open-vocabulary and VLM/LLM-based 3D scene graph approaches, alongside dynamic 3D Gaussian Splatting representations \cite{Wang2025GaussianGraph3G}, further study generalization beyond closed predicate sets and view-invariant graph construction~\cite{koch2024open3dsg,gu2024conceptgraphs,Chen2024CLIPDrivenO3,madhavaram2026vizor}.

% Despite these advances, existing supervised closed-vocabulary 3DSGG methods largely overlook predicate-level transformation behavior under viewpoint changes.
% They typically predict all predicates in the original benchmark label space without explicitly distinguishing relations that should remain stable from those that should transform under yaw viewpoint changes.
% We identify this predicate-level transformation heterogeneity as a key factor for viewpoint-robust 3DSGG and address it with a transformation-aware relation modeling framework that decouples yaw-invariant and direction-sensitive relation cues while preserving unified 26-class predicate prediction.

% Despite these advances, existing 3DSGG methods largely overlook predicate-level transformation heterogeneity under yaw viewpoint changes, and typically predict all predicates in a shared benchmark label space without distinguishing relations that should remain stable from those that should transform. We identify this heterogeneity as a key factor for viewpoint-robust 3DSGG and address it with a transformation-aware relation modeling framework that decouples yaw-invariant and direction-sensitive relation cues while preserving unified multi-label predicate prediction.

Despite these advances, existing 3DSGG methods largely overlook the heterogeneous dependency of relation semantics on the observer frame. They typically predict all predicates in a unified label space without distinguishing relations that should remain stable from those that should be reinterpreted under observer-frame reorientation. We address this issue with observer-aware relation modeling that captures Observer-Independent and Observer-Dependent cues while preserving unified multi-label predicate prediction.

% Rework的第二段和第三段合并，只写两段。第一段泛讲3DSGG，第二段应该主要讲realtion建模，因为你这篇的重心是rotation robust的relation modeling，所以第二段的关键词应该是：第一段：current relation modeling（回顾相关工作，一般是近两三年） + directional invariant representation learning （讲讲其他领域怎么处理这个问题的，你的思路是从哪来的，而这个问题在sgg没被重视）；第二段：与以往工作不同的是，我们怎么把这种思路引入到3DSGG从而实现方向鲁棒地关系预测。rework修改完毕后长度应该大概在第三页左侧中下，先不要超过第三页左侧。
\subsection{Geometry-Aware Scene Understanding}

Relation modeling is central to scene graph generation.
In 2DSGG, prior work has studied contextualized relation prediction, debiasing, bipartite message passing, and knowledge-guided reasoning to address long-tailed predicates and semantic ambiguity~\cite{zellers2018neural,tang2019learning,tang2020unbiased,li2021bipartite,zareian2020bridging,chen2019knowledge}.
In 3DSGG, predicate prediction is commonly formulated as multi-label classification over directed object pairs, where object features are combined with edge-specific geometric descriptors such as displacement, distance, orientation, and scale ratios~\cite{wald2020learning,zhang2021exploiting,wang2023vl}.
Meanwhile, rotation-equivariant and rotation-invariant representations have been widely explored in geometric deep learning~\cite{bronstein2021geometric}, including group equivariant networks over $E(2)$ and $SE(3)$~\cite{Cohen2016GroupEC,Weiler2019GeneralES,Thomas2018TensorFN,fuchs2020se,Yao2024PARENetPR}, $SO(3)$-equivariant point-cloud models such as Vector Neurons~\cite{deng2021vector}, and rotation-invariant representations or descriptors~\cite{su2025ri,rao2019spherical}.

% Importantly, our goal is not to enforce full $SO(3)$ invariance or equivariance for all predicates.
% In gravity-aligned 3DSGG, such a uniform rotation constraint can be overly restrictive: directional predicates should change with the yaw frame, whereas most other predicates should remain stable.
% Existing supervised 3DSGG methods still predict all predicates in a shared closed-vocabulary relation space, without separating relations that transform differently under yaw viewpoint changes.
% TAD addresses this gap by decoupling relation reasoning with transformation-specific descriptors, parameter-independent branches, and group-aware supervision, while preserving unified 26-class multi-label prediction.

% These works inspire transformation-aware representation design, but mainly focus on point-level or object-level stability rather than predicate-level relation behavior. Instead of enforcing a uniform $SO(3)$ invariance or equivariance constraint for all predicates, we focus on gravity-aligned 3DSGG, where directional predicates should change with the yaw frame while most other predicates should remain stable. TAD addresses this gap by decoupling relation reasoning with transformation-specific descriptors, parameter-decoupled branches, and group-aware supervision, while preserving unified  multi-label prediction.

These works motivate geometry-aware representation design, but mainly focus on point-level or object-level stability rather than predicate-specific observer-frame dependency. 
% Instead of imposing a uniform invariance or equivariance constraint on all predicates, we distinguish relation cues according to their different frame responses. 
In gravity-aligned 3DSGG, relation prediction does not require imposing a uniform invariance or equivariance constraint on all predicates: directional predicates should be reinterpreted with the observer frame, whereas most contact, support, and semantic predicates should remain stable.
In contrast to methods that impose a uniform geometric treatment on all predicates, OAR combines observer-aware geometric encoding and relation specialization to account for heterogeneous observer-frame dependencies, supported by frame-stable object encoding, while retaining unified multi-label prediction.
\section{Method}
\label{sec:method}

\subsection{Problem Formulation}
\label{sec:problem_formulation}
\label{sec:predicate_transformation}

Given a 3D point cloud $\mathbf{P}\in\mathbb{R}^{N\times 3}$ and
class-agnostic instance masks
$\mathcal{M}=\{M_i\}_{i=1}^{K}$,
the conventional 3DSGG task aims to predict a directed 3D scene graph
$\mathcal{G}=(\mathcal{V},\mathcal{E})$.
Each node $v_i\in\mathcal{V}$ denotes an object instance with label $o_i$,
and each directed edge $e_{ij}\in\mathcal{E}$ encodes predicates from
subject $i$ to object $j$.
Let $\mathcal{C}_{\mathrm{rel}}$ denote the predicate vocabulary and
$N_{\mathrm{rel}}=|\mathcal{C}_{\mathrm{rel}}|$ its cardinality.
The model predicts object logits $\hat{\mathbf{o}}_i$ and predicate logits
$\mathbf{s}_{ij}\in\mathbb{R}^{N_{\mathrm{rel}}}$.
Since 3DSSG relations are multi-label, each edge is supervised by a
multi-hot vector
$\mathbf{y}_{ij}\in\{0,1\}^{N_{\mathrm{rel}}}$.

To study relation prediction under observer-frame reorientation, we re-express the same physical scene in different horizontal observer frames while keeping the gravity-aligned vertical axis fixed. The object instances, masks, and visibility remain unchanged. For a controlled diagnosis, our main evaluation uses the cardinal subset $\Theta_{\mathrm{card}} =\{0^\circ,90^\circ,180^\circ,270^\circ\}$. At these orientations, the horizontal axes map exactly onto one another, so directional predicate labels admit deterministic re-indexing without boundary ambiguity. This subset defines the evaluation protocol only; the model operates on continuous geometric features and is not architecturally restricted to cardinal angles.

Relations exhibit different dependencies on the observer frame. We define $\mathcal{C}_{\mathrm{dep}} =\{\textit{left},\textit{front},\textit{right},\textit{behind}\}$ as \textbf{Observer-Dependent Relations}, whose labels change with the horizontal observer frame. The remaining predicates form the approximately \textbf{Observer-Independent Relations}, $\mathcal{C}_{\mathrm{ind}} =\mathcal{C}_{\mathrm{rel}}\setminus\mathcal{C}_{\mathrm{dep}}$, whose labels remain unchanged. 
For $\theta\in\Theta_{\mathrm{card}}$, the expected mapping is $T_{\theta}(r)=\Pi_{\theta}(r)$ for $r\in\mathcal{C}_{\mathrm{dep}}$ and $T_{\theta}(r)=r$ otherwise, where $\Pi_{\theta}$ is the corresponding deterministic predicate permutation.

The four cardinal rotations form the cyclic group $C_4$.
Its $180^\circ$ element generates the subgroup
$C_2=\{0^\circ,180^\circ\}\subset C_4$, under which directions are only
reversed within the left--right and front--behind axes.
In contrast, the $90^\circ$ and $270^\circ$ elements exchange these two
axes.
We use this subgroup structure as a diagnostic hierarchy:
the $C_2$ case probes within-axis reversal, whereas the remaining elements of
$C_4$ additionally probe cross-axis reinterpretation.
This distinction characterizes different observer-frame reasoning demands
without imposing a strict group-equivariant constraint on the model.

\begin{figure}[t]
\centering
\includegraphics[width=\columnwidth]{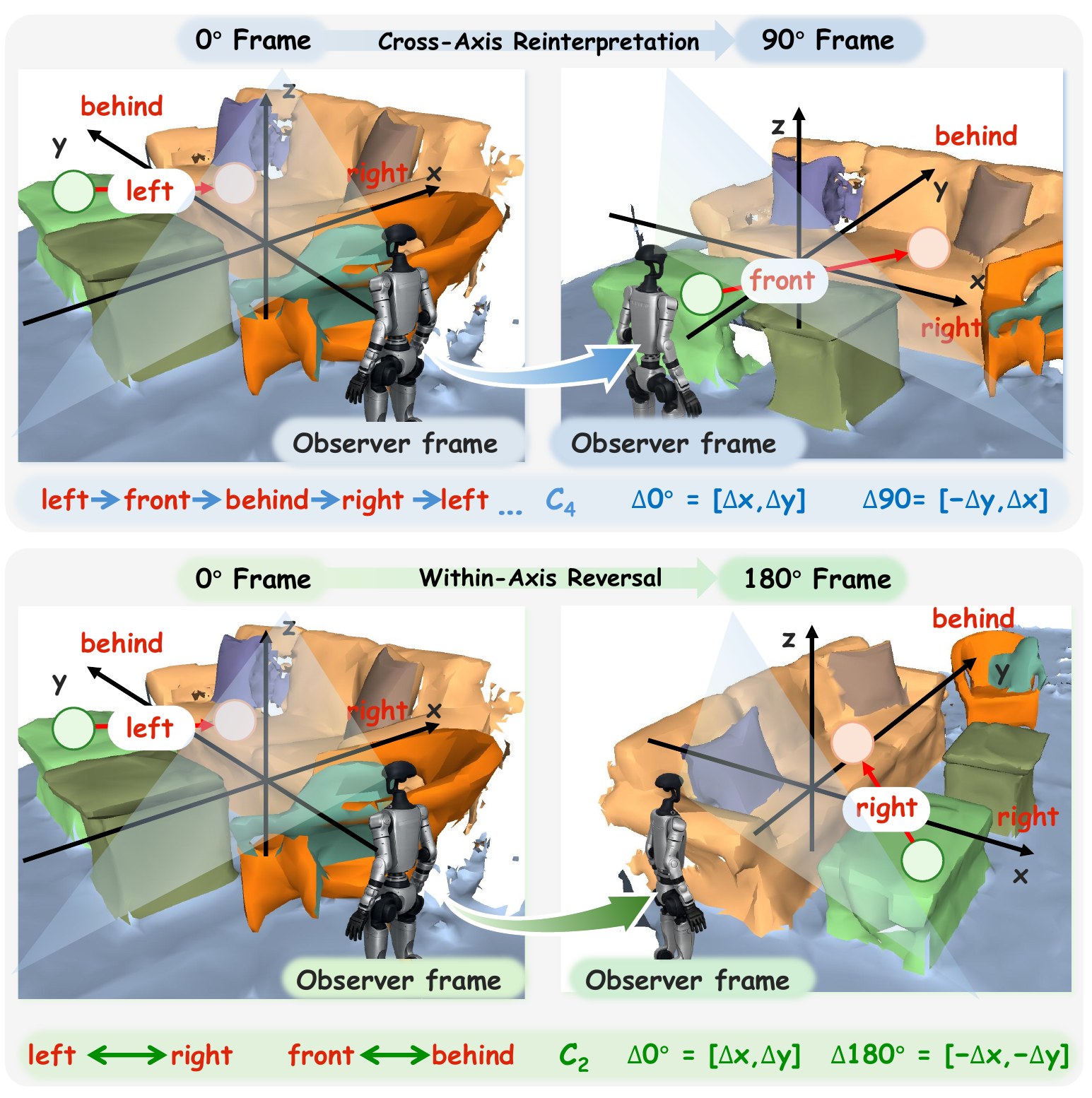}
\caption{
Representative cardinal observer-frame reorientations. The $0^\circ\!\rightarrow\!90^\circ$ case exchanges the horizontal axes and probes cross-axis reinterpretation, whereas the $0^\circ\!\rightarrow\!180^\circ$ case reverses directions within each axis and generates the subgroup $C_2=\{0^\circ,180^\circ\}\subset C_4$.
}
\label{fig:prelim_axis_exchange}
\end{figure}

As illustrated in Fig.~\ref{fig:prelim_axis_exchange}, correctly handling opposite-direction reversal under $C_2$ does not necessarily imply reliable reinterpretation across the left--right and front--behind axes. The full derivation, predicate permutation table, and categorization of $\mathcal{C}_{\mathrm{rel}}$ are provided in the supplementary material. This heterogeneous observer-frame dependency motivates relation reasoning that models observer-independent and observer-dependent cues according to their distinct frame responses while preserving the unified predicate space.

\begin{figure*}[t]
\centering
\includegraphics[width=\textwidth]{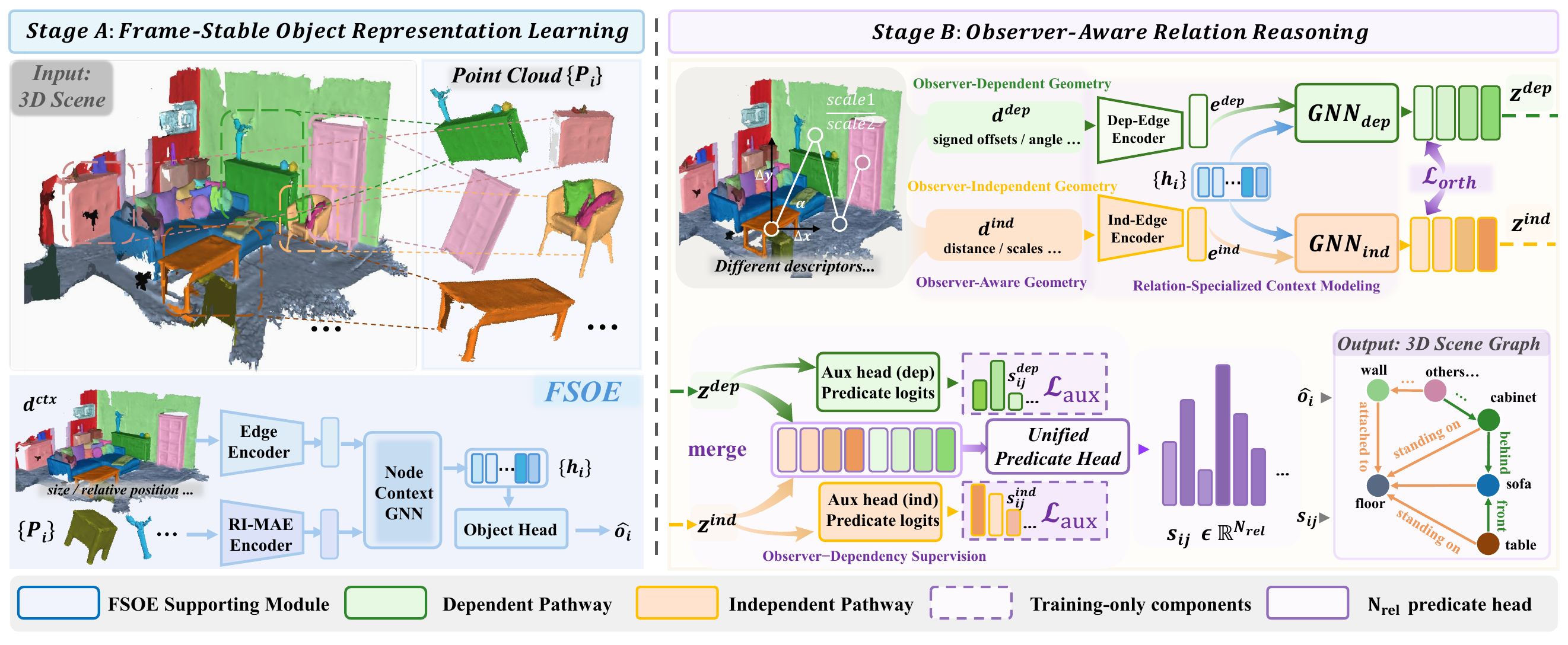}
\caption{
Overview of OAR.
During training, FSOE extracts rotation-stable object features, and the core Observer-Aware Relation Reasoning specializes observer-independent and observer-dependent relation cues with auxiliary supervision.
During inference, auxiliary heads are removed, and the merged pathway representations predict predicate logits over $\mathcal{C}_{\mathrm{rel}}$.
}
\label{fig:pipeline}
\end{figure*}

\subsection{Architecture}
\label{sec:overall_framework}

% As shown in Fig.~\ref{fig:pipeline}, the Frame-Stable Object Encoder (FSOE) first extracts frame-stable object features from segmented point clouds and then incorporates scene context to produce object representations $\{\mathbf{h}_i\}_{i=1}^{K}$, from which object logits are predicted. For each directed pair $(i,j)$, OAR produces observer-independent and observer-dependent representations $\mathbf{z}_{ij}^{\mathrm{ind}}$ and $\mathbf{z}_{ij}^{\mathrm{dep}}$, which are merged to predict multi-label predicate logits
% $\mathbf{s}_{ij}$ over $\mathcal{C}_{\mathrm{rel}}$.

As shown in Fig.~\ref{fig:pipeline}, OAR follows a two-stage pipeline. The \textbf{Frame-Stable Object Encoder} (FSOE) extracts rotation-stable object features and incorporates scene context to produce object representations $\{\mathbf{h}_i\}_{i=1}^{K}$.
For each directed pair $(i,j)$, \textbf{Observer-Aware Geometric Encoding} (OAGE) constructs observer-independent and observer-dependent geometric descriptors. The two cues are further specialized through non-shared relation GNNs and observer-dependency supervision, producing representations $\mathbf{z}_{ij}^{\mathrm{ind}}$ and $\mathbf{z}_{ij}^{\mathrm{dep}}$. The representations are merged to predict multi-label predicate logits $\mathbf{s}_{ij}$ over $\mathcal{C}_{\mathrm{rel}}$.

\subsection{Frame-Stable Object Encoder}
\label{sec:vsoe}
Given the object point cloud $\mathbf{P}_i$ of instance $i$, FSOE extracts a rotation-stable object feature using a pretrained RI-MAE encoder~\cite{su2025ri}:
\begin{equation}
\mathbf{z}_i^0
=
W_p f_{\mathrm{RI\text{-}MAE}}(\mathbf{P}_i),
\end{equation}
where $W_p$ projects the RI-MAE feature to the graph reasoning dimension. A node-level GNN then incorporates lightweight contextual edge descriptors to produce scene-aware object representations $\mathbf{h}_i$, from which object logits are predicted as $\hat{\mathbf{o}}_i=f_{\mathrm{obj}}(\mathbf{h}_i)$.

\subsection{Observer-Aware Relation Reasoning}
\label{sec:tad}

The relation reasoning stage models observer-independent and observer-dependent relation cues according to their distinct responses to observer-frame reorientation. It consists of observer-aware geometric encoding and a relation-specialization design that combines non-shared context modeling with observer-dependency supervision.

\noindent\textbf{Observer-Aware Geometric Encoding.}
For each directed object pair $(i,j)$, where $i$ and $j$ denote the subject and object, respectively, we define the signed centroid difference as $\Delta\boldsymbol{\mu}_{ij} =\boldsymbol{\mu}_{i}-\boldsymbol{\mu}_{j} =[\Delta x_{ij},\Delta y_{ij},\Delta z_{ij}]$, the horizontal distance $\rho_{ij}=\sqrt{\Delta x_{ij}^{2}+\Delta y_{ij}^{2}}$, the 3D distance $d^{3D}_{ij}=\|\Delta\boldsymbol{\mu}_{ij}\|_{2}$, and the horizontal direction angle $\varphi_{ij}=\mathrm{atan2}(\Delta y_{ij},\Delta x_{ij})$.

The observer-dependent descriptor preserves signed offsets and angular cues:
\begin{equation}
\mathbf{d}^{\mathrm{dep}}_{ij}
=
[
\Delta x_{ij}, \Delta y_{ij},
\cos\varphi_{ij}, \sin\varphi_{ij},
\Delta z_{ij},
r^{v}_{ij}, r^{l}_{ij}
].
\label{eq:dir_descriptor}
\end{equation}

The observer-independent descriptor instead suppresses signed horizontal
orientation:
\begin{equation}
\mathbf{d}^{\mathrm{ind}}_{ij}
=
[
\rho_{ij}, d^{3D}_{ij}, \Delta z_{ij},
r^{v}_{ij}, r^{l}_{ij},
\boldsymbol{\eta}_{i},
\boldsymbol{\eta}_{j}
].
\label{eq:inv_descriptor}
\end{equation}
Here, $r^{v}_{ij}$ and $r^{l}_{ij}$ are log-scale ratios, while $\boldsymbol{\eta}_{i}$ and $\boldsymbol{\eta}_{j}$ denote object shape and scale statistics.
% Thus, $\mathbf{d}^{\mathrm{dep}}_{ij}$ retains signed horizontal quantities that vary with the observer frame, whereas $\mathbf{d}^{\mathrm{ind}}_{ij}$ represents horizontal geometry through orientation-free magnitudes.
Detailed definitions are provided in the supplementary material.

\noindent\textbf{Relation-Specialized Context Modeling.}
% Each descriptor is projected into a pathway-specific edge feature $\mathbf{e}^{b,0}_{ij}=\phi_b(\mathbf{d}^{b}_{ij})$, where $b\in\{\mathrm{ind},\mathrm{dep}\}$. The relation reasoning stage then applies two non-shared relation GNNs: $\mathbf{z}^{b}_{ij} =\mathrm{GNN}_{b} (\mathbf{h}_{i},\mathbf{h}_{j},\mathbf{e}^{b,0}_{ij})$. The non-shared parameters allow each pathway to specialize to its corresponding observer-frame dependency.
Each descriptor is projected into a pathway-specific edge feature $\mathbf{e}^{b,0}_{ij}=\phi_b(\mathbf{d}^{b}_{ij})$, where $b\in\{\mathrm{ind},\mathrm{dep}\}$.
Both pathways operate on the same directed object graph and share the frame-stable node representations $\mathbf{h}_i$ and $\mathbf{h}_j$, while using different geometric descriptors and non-shared relation-GNN parameters.
The relation reasoning stage is formulated as $\mathbf{z}^{b}_{ij} =\mathrm{GNN}_{b}(\mathbf{h}_{i},\mathbf{h}_{j},\mathbf{e}^{b,0}_{ij})$.
The non-shared parameters allow the observer-independent pathway to emphasize stable structural and contextual cues, whereas the observer-dependent pathway preserves signed directional information required for observer-frame-sensitive relations.
The two pathway representations remain separate during relation reasoning and are fused only before final predicate prediction.

\noindent\textbf{Observer-Dependency Supervision.} 
% To encourage complementary relation representations, we use a feature-level orthogonal regularizer: \begin{equation} \mathcal{L}_{orth} = \frac{1}{|\mathcal{E}|} \sum_{(i,j)\in\mathcal{E}} \left| \frac{ \langle \mathbf{z}^{\mathrm{ind}}_{ij}, \mathbf{z}^{\mathrm{dep}}_{ij} \rangle }{ \|\mathbf{z}^{\mathrm{ind}}_{ij}\|_2 \|\mathbf{z}^{\mathrm{dep}}_{ij}\|_2 + \epsilon } \right|. \label{eq:orth_loss} \end{equation}
% This regularizer reduces redundancy between two pathway representations without imposing mutual exclusivity, since relations from both subsets may co-occur on the same edge.
% During training, auxiliary heads predict $\mathbf{s}^{\mathrm{ind}}_{ij} =f^{\mathrm{ind}}_{\mathrm{aux}}(\mathbf{z}^{\mathrm{ind}}_{ij})$ and $\mathbf{s}^{\mathrm{dep}}_{ij} =f^{\mathrm{dep}}_{\mathrm{aux}}(\mathbf{z}^{\mathrm{dep}}_{ij})$ over $\mathcal{C}_{\mathrm{ind}}$ and $\mathcal{C}_{\mathrm{dep}}$, respectively, without assuming mutual exclusivity. The auxiliary loss is $\mathcal{L}_{aux} =\mathcal{L}^{\mathrm{ind}}_{aux} +\beta\mathcal{L}^{\mathrm{dep}}_{aux}$, where both terms use binary cross-entropy. The auxiliary heads are removed during inference. Together with the non-shared pathways, this supervision forms our relation-specialization design.
Although non-shared parameters enable pathway specialization, they do not guarantee complementary relation representations.
We therefore use a feature-level orthogonal regularizer:
\begin{equation}
\mathcal{L}_{orth}
=
\frac{1}{|\mathcal{E}|}
\sum_{(i,j)\in\mathcal{E}}
\left|
\frac{
\langle
\mathbf{z}^{\mathrm{ind}}_{ij},
\mathbf{z}^{\mathrm{dep}}_{ij}
\rangle
}{
\|\mathbf{z}^{\mathrm{ind}}_{ij}\|_2
\|\mathbf{z}^{\mathrm{dep}}_{ij}\|_2
+
\epsilon
}
\right|.
\label{eq:orth_loss}
\end{equation}
This regularizer reduces redundancy between the two pathway representations without enforcing a hard separation, since Observer-Independent and Observer-Dependent Relations may co-occur on the same edge.

During training, auxiliary heads predict $\mathbf{s}^{\mathrm{ind}}_{ij} =f^{\mathrm{ind}}_{\mathrm{aux}}(\mathbf{z}^{\mathrm{ind}}_{ij})$ and $\mathbf{s}^{\mathrm{dep}}_{ij} =f^{\mathrm{dep}}_{\mathrm{aux}}(\mathbf{z}^{\mathrm{dep}}_{ij})$ over $\mathcal{C}_{\mathrm{ind}}$ and $\mathcal{C}_{\mathrm{dep}}$, respectively.
Their supervision targets are obtained from the corresponding dimensions of the original multi-hot relation label, rather than by assigning each edge exclusively to one pathway.
Therefore, an edge containing relations from both subsets contributes supervision to both auxiliary heads.
The auxiliary loss is $\mathcal{L}_{aux} = \mathcal{L}^{\mathrm{ind}}_{aux} + \beta\mathcal{L}^{\mathrm{dep}}_{aux}$, where both terms use binary cross-entropy. The auxiliary heads encourage semantic specialization of the two pathways, while $\mathcal{L}_{orth}$ reduces redundancy between their feature representations. They are used only during training and removed during inference. Together with the non-shared relation GNNs, these two forms of supervision constitute our relation-specialization design.

\noindent\textbf{Unified Predicate Prediction.} 
%Although the two relation cues are modeled through specialized pathways, final prediction is performed over the original unified predicate space: \begin{equation} \mathbf{s}_{ij} = f_{\mathrm{cls}} \left( f_{\mathrm{merge}} \left( [ \mathbf{z}^{\mathrm{ind}}_{ij}, \mathbf{z}^{\mathrm{dep}}_{ij} ] \right) \right), \qquad \mathbf{s}_{ij}\in\mathbb{R}^{N_{\mathrm{rel}}}. \label{eq:predicate_logits} \end{equation} This design specializes relation reasoning without imposing a hard partition on the final predictions, allowing relations from both subsets to co-occur on the same edge.
Although the two relation cues are modeled through specialized pathways, final prediction is performed over the original unified predicate space:
\begin{equation}
\mathbf{s}_{ij}
=
f_{\mathrm{cls}}
\left(
f_{\mathrm{merge}}
\left(
[
\mathbf{z}^{\mathrm{ind}}_{ij},
\mathbf{z}^{\mathrm{dep}}_{ij}
]
\right)
\right),
\qquad
\mathbf{s}_{ij}\in\mathbb{R}^{N_{\mathrm{rel}}}.
\label{eq:predicate_logits}
\end{equation}
This design specializes relation reasoning without imposing a hard partition on the final predictions, allowing relations from both subsets to co-occur on the same edge.

\subsection{Loss Functions}
\label{sec:training_objective}

The training objective combines object classification and relation prediction.
For object classification, we use
$\mathcal{L}_{obj}
=\frac{1}{K}\sum_{i=1}^{K}
\mathrm{CE}(\hat{\mathbf{o}}_i,o_i)$.
For relation prediction, we apply binary cross-entropy to the unified
predicate logits:
$\mathcal{L}_{bce}
=\frac{1}{|\mathcal{E}|}
\sum_{(i,j)\in\mathcal{E}}
\mathrm{BCE}(\mathbf{s}_{ij},\mathbf{y}_{ij})$.
The relation loss is
\begin{equation}
\mathcal{L}_{rel}
=
\mathcal{L}_{bce}
+
\lambda_{orth}\mathcal{L}_{orth}
+
\lambda_{aux}\mathcal{L}_{aux}.
\label{eq:rel_loss}
\end{equation}
The overall objective is
$\mathcal{L}
=
\lambda_{obj}\mathcal{L}_{obj}
+
\lambda_{rel}\mathcal{L}_{rel}$.
\section{Experiments}
\label{sec:experiments}

\subsection{Experimental Setup}
\label{sec:exp_setup}

\paragraph{Dataset.}
We conduct experiments on the 3DSSG dataset~\cite{wald2020learning}, a scene graph benchmark built upon 3RScan~\cite{wald2019rio} with 3D reconstructed indoor scenes. Following the standard protocol, we use the official train/validation split. The benchmark contains 160 object categories and 26 predicate categories. We treat \textit{left}, \textit{front}, \textit{right}, and \textit{behind} as Observer-Dependent Relations ($\mathcal{C}_{\mathrm{dep}}$), and the remaining 22 predicates as approximately Observer-Independent Relations ($\mathcal{C}_{\mathrm{ind}}$) under horizontal observer-frame reorientation.

%\paragraph{Evaluation metrics.}
% For standard 3DSGG, we report object (R@1, R@5), predicate (R@1, R@3), and triplet (R@50, R@100) prediction, as well as SGCls and PredCls with recall and mean recall. 
% For viewpoint robustness, we adopt a controlled protocol: the physical scene is kept fixed, while its input coordinates are expressed in yaw-rotated observation frames with $\theta\in\{0^\circ,90^\circ,180^\circ,270^\circ\}$. We apply the corresponding deterministic permutation to observation-frame-dependent directional labels. The complete permutation table and empirical label-consistency validation are provided in the supplementary material; GT directional labels show 99.38\% sign consistency with observation-frame displacement, increasing to 99.96--100.00\% after removing boundary-ambiguous cases.

\begin{table}[t]
\centering
\caption{Comparison on standard 3DSGG. Best in \textbf{bold}, second-best \underline{underlined}.}
\label{tab:comparison_results}
\footnotesize
\setlength{\tabcolsep}{3.5pt}
\resizebox{\columnwidth}{!}{%
\begin{tabular}{lcccccc}
\toprule
\multirow{2}{*}{Model}
& \multicolumn{2}{c}{Object}
& \multicolumn{2}{c}{Predicate}
& \multicolumn{2}{c}{Triplet} \\
\cmidrule(lr){2-3}
\cmidrule(lr){4-5}
\cmidrule(lr){6-7}
& R@1 & R@5
& R@1 & R@3
& R@50 & R@100 \\
\midrule
SGPN \cite{wang2018sgpn}
& 49.46 & 73.99
& 86.92 & 94.76
& 85.38 & 88.59 \\

SGFN \cite{wu2021scenegraphfusion}
& 53.36 & 76.88
& 89.00 & 97.71
& 88.59 & 91.14 \\

VL-SAT \cite{wang2023vl}
& 55.93 & 78.06
& 89.81 & \underline{98.46}
& 89.35 & 92.20 \\

VL-SAT+RI-MAE
& \underline{59.41} & \underline{79.81}
& 90.32 & 96.70
& 90.26 & 93.16 \\

OCRL \cite{heo2025object}
& 59.13 & 79.20
& \textbf{91.72} & \textbf{98.48}
& \textbf{91.40} & \underline{93.80} \\

\midrule
Ours
& \textbf{60.51} & \textbf{81.79}
& \underline{91.36} & 98.36
& \underline{91.03} & \textbf{93.84} \\
\bottomrule
\end{tabular}%
}
\end{table}

% We report Overall R@50, directional mR@50, and invariant mR@50 under PredCls with ground-truth object labels as the robustness protocol, since it isolates predicate-level transformation behavior from confounding object-classification errors. SGCls rotation results are reported as complementary evidence in the supplementary material.
\paragraph{Evaluation metrics.}
% For standard 3DSGG, we report object (R@1, R@5), predicate (R@1, R@3), and triplet (R@50, R@100) prediction, together with recall and mean recall for SGCls and PredCls. For observer-frame reorientation, we follow the controlled protocol in Problem Formulation: the physical scene, instances, masks, and visibility remain fixed, while the complete scene is re-expressed under $\Theta_{\mathrm{card}} =\{0^\circ,90^\circ,180^\circ,270^\circ\}$. Observer-Dependent Relation labels are re-indexed by the corresponding permutation. The permutation table and label-consistency analysis are provided in the supplementary material; Observer-Dependent label consistency increases from 99.38\% to 99.96--100.00\% after excluding boundary-ambiguous cases. We also report evaluations at $30^\circ$ intervals in the supplementary material.

% We report Overall R@50, observer-dependent mR@50, and observer-independent mR@50 under PredCls with ground-truth object labels, isolating predicate-level observer-frame responses. Complementary SGCls results are provided in the supplementary material.
For the standard 3DSGG setting, we report object R@1/R@5, predicate R@1/R@3, and triplet R@50/R@100, together with recall and mean recall for SGCls and PredCls. For observer-frame evaluation, we follow the cardinal reorientation protocol defined in Problem Formulation and re-index Observer-Dependent labels using the corresponding predicate permutation. The permutation table and label-consistency analysis are provided in the supplementary material; after excluding boundary-ambiguous cases, the consistency of re-indexed Observer-Dependent labels increases from 99.38\% to 99.96--100.00\%. Unless otherwise stated, we report PredCls Overall R@50, Observer-Dependent mR@50, and Observer-Independent mR@50, thereby isolating predicate-level frame responses from object recognition errors. SGCls and dense-angle results are provided in the supplementary material.

\paragraph{Baselines.}
% We compare with SGPN~\cite{wang2018sgpn}, SGFN~\cite{wu2021scenegraphfusion}, VL-SAT~\cite{wang2023vl}, Zhang et al.~\cite{zhang2021knowledge}, and OCRL~\cite{heo2025object}. We also include rotation-augmented variants ``(aug)'', where each training scene is expanded into four yaw-rotated copies at $\{0^\circ,90^\circ,180^\circ,270^\circ\}$, with directional predicate labels transformed by the corresponding permutation. This quadruples the downstream training set and tests whether conventional architectures can learn cardinal yaw transformations from rotation-augmented data alone. For these augmented baselines, we use the same optimizer, loss weights, and validation-based checkpoint selection as their non-augmented counterparts, and allow extended training until validation performance saturates, avoiding under-training the augmented models. VL-SAT+RI-MAE uses the same RI-MAE encoder as our method for controlled comparison.
We compare with SGPN~\cite{wang2018sgpn}, SGFN~\cite{wu2021scenegraphfusion}, VL-SAT~\cite{wang2023vl}, Zhang et al.~\cite{zhang2021knowledge}, and OCRL~\cite{heo2025object}. We also include cardinal frame-reorientation-augmented variants ``(aug)'', where each training scene is expanded into four copies at $\{0^\circ,90^\circ,180^\circ,270^\circ\}$, with Observer-Dependent Relation labels re-indexed by the corresponding permutation. This quadruples the downstream training set and tests whether conventional architectures can learn frame-induced predicate re-indexing through augmentation alone. For augmented baselines, we use the same optimizer, loss weights, and validation-based checkpoint selection as their non-augmented counterparts, and train until validation performance saturates. VL-SAT+RI-MAE uses the same RI-MAE encoder as our method for controlled comparison.

\begin{table*}[t]
\centering
\caption{Standard SGCls and PredCls evaluation on 3DSSG. Best in \textbf{bold}, second-best \underline{underlined}.}
\label{tab:sgcls_predcls}
\small
\setlength{\tabcolsep}{6pt}
\begin{tabular}{lcccc}
\toprule
\multirow{2}{*}{Method}
& \multicolumn{2}{c}{SGCls}
& \multicolumn{2}{c}{PredCls} \\
\cmidrule(lr){2-3}
\cmidrule(lr){4-5}
& R@\{20/50/100\}
& mR@\{20/50/100\}
& R@\{20/50/100\}
& mR@\{20/50/100\} \\
\midrule
SGPN \cite{wang2018sgpn}
& 27.0/28.8/29.0
& 19.7/22.6/23.1
& 51.9/58.0/58.5
& 32.1/38.4/38.9 \\

Zhang et al. \cite{zhang2021knowledge}
& 28.5/30.0/30.1
& 24.4/28.6/28.8
& 59.3/65.0/65.3
& 56.6/63.5/63.8 \\

SGFN \cite{wu2021scenegraphfusion}
& 29.5/31.2/31.2
& 20.5/23.1/23.1
& 65.9/78.8/79.6
& 46.1/54.8/55.1 \\

VL-SAT \cite{wang2023vl}
& 32.0/33.5/33.7
& \underline{31.0}/\underline{32.6}/\underline{32.7}
& 67.8/79.9/80.8
& \underline{57.8}/64.2/64.3 \\

OCRL \cite{heo2025object}
& \underline{36.1}/\underline{37.7}/\underline{37.8}
& 29.8/32.0/32.1
& \underline{70.2}/\underline{82.0}/\underline{82.6}
& 57.1/\underline{64.6}/\underline{64.8} \\

\textbf{Ours}
& \textbf{36.4}/\textbf{38.1}/\textbf{38.3}
& \textbf{32.0}/\textbf{34.6}/\textbf{34.7}
& \textbf{73.1}/\textbf{86.0}/\textbf{86.4}
& \textbf{58.8}/\textbf{66.7}/\textbf{67.1} \\
\bottomrule
\end{tabular}
\end{table*}

\begin{table*}[t]
  \centering
  \caption{
%   Relation-focused robustness under yaw rotations (PredCls, ground-truth object labels). Gray $\Delta$ columns report change relative to $0^\circ$. ``(aug)'' denotes training-time rotation augmentation using four yaw-rotated copies per training scene with corresponding predicate label permutations.
Relation-focused evaluation under cardinal observer-frame reorientation (PredCls with ground-truth object labels). Gray $\Delta$ columns report changes relative to the $0^\circ$ reference frame. ``(aug)'' denotes cardinal frame-reorientation augmentation using four re-expressed copies per training scene with the corresponding Observer-Dependent Relation label permutations.}
  \label{tab:rotated_view_robustness}
  \footnotesize
  \setlength{\tabcolsep}{1.5pt}
  \renewcommand{\arraystretch}{1.15}
  \resizebox{\textwidth}{!}{%
  \begin{tabular}{lccccccccccccccccccccc}
    \toprule
    \multirow{2}{*}{Model}
    & \multicolumn{7}{c}{Overall R@50}
    & \multicolumn{7}{c}{Observer-Dependent mR@50}
    & \multicolumn{7}{c}{Observer-Independent mR@50} \\
    \cmidrule(lr){2-8}
    \cmidrule(lr){9-15}
    \cmidrule(lr){16-22}
    & $0^\circ$
    & $90^\circ$ & $\Delta_{90^\circ}$
    & $180^\circ$ & $\Delta_{180^\circ}$
    & $270^\circ$ & $\Delta_{270^\circ}$
    & $0^\circ$
    & $90^\circ$ & $\Delta_{90^\circ}$
    & $180^\circ$ & $\Delta_{180^\circ}$
    & $270^\circ$ & $\Delta_{270^\circ}$
    & $0^\circ$
    & $90^\circ$ & $\Delta_{90^\circ}$
    & $180^\circ$ & $\Delta_{180^\circ}$
    & $270^\circ$ & $\Delta_{270^\circ}$ \\
    \midrule

    VL-SAT \cite{wang2023vl}
      & 79.9
      & 68.2 & \cellcolor{gray!15}$-11.7$
      & 79.6 & \cellcolor{gray!15}$-0.3$
      & 69.3 & \cellcolor{gray!15}$-10.6$
      & 82.1
      & 67.8 & \cellcolor{gray!15}$-14.3$
      & 82.3 & \cellcolor{gray!15}$+0.2$
      & 68.4 & \cellcolor{gray!15}$-13.7$
      & 60.9
      & 51.1 & \cellcolor{gray!15}$-9.8$
      & 60.4 & \cellcolor{gray!15}$-0.5$
      & 50.8 & \cellcolor{gray!15}$-10.1$ \\

    VL-SAT+RI-MAE
      & 81.9
      & 71.0 & \cellcolor{gray!15}$-10.9$
      & 81.5 & \cellcolor{gray!15}$-0.4$
      & 71.8 & \cellcolor{gray!15}$-10.1$
      & 83.7
      & 70.3 & \cellcolor{gray!15}$-13.4$
      & 83.5 & \cellcolor{gray!15}$-0.2$
      & 70.1 & \cellcolor{gray!15}$-13.6$
      & 61.2
      & 54.7 & \cellcolor{gray!15}$-6.5$
      & 61.6 & \cellcolor{gray!15}$+0.4$
      & 55.1 & \cellcolor{gray!15}$-6.1$ \\

    VL-SAT$_{(aug)}$
      & 71.9
      & 73.2 & \cellcolor{gray!15}$+1.3$
      & 72.4 & \cellcolor{gray!15}$+0.5$
      & 72.2 & \cellcolor{gray!15}$+0.3$
      & 81.8
      & 79.3 & \cellcolor{gray!15}$-2.5$
      & 82.1 & \cellcolor{gray!15}$+0.3$
      & 78.9 & \cellcolor{gray!15}$-2.9$
      & 56.2
      & 56.1 & \cellcolor{gray!15}$-0.1$
      & 55.6 & \cellcolor{gray!15}$-0.6$
      & 55.9 & \cellcolor{gray!15}$-0.3$ \\

    OCRL \cite{heo2025object}
      & 82.0
      & 69.9 & \cellcolor{gray!15}$-12.1$
      & 81.8 & \cellcolor{gray!15}$-0.2$
      & 71.6 & \cellcolor{gray!15}$-10.4$
      & 83.9
      & 67.4 & \cellcolor{gray!15}$-16.5$
      & 84.5 & \cellcolor{gray!15}$+0.6$
      & 69.1 & \cellcolor{gray!15}$-14.8$
      & 61.1
      & 53.1 & \cellcolor{gray!15}$-8.0$
      & 60.9 & \cellcolor{gray!15}$-0.2$
      & 53.5 & \cellcolor{gray!15}$-7.6$ \\

    OCRL$_{(aug)}$
      & 76.8
      & 76.0 & \cellcolor{gray!15}$-0.8$
      & 76.1 & \cellcolor{gray!15}$-0.7$
      & 75.9 & \cellcolor{gray!15}$-0.9$
      & 80.9
      & 78.0 & \cellcolor{gray!15}$-2.9$
      & 81.2 & \cellcolor{gray!15}$+0.3$
      & 77.4 & \cellcolor{gray!15}$-3.5$
      & 57.4
      & 57.7 & \cellcolor{gray!15}$+0.3$
      & 56.8 & \cellcolor{gray!15}$-0.6$
      & 57.9 & \cellcolor{gray!15}$+0.5$ \\

    \textbf{Ours}
      & \textbf{86.0}
      & \textbf{84.5} & \cellcolor{gray!15}$-1.5$
      & \textbf{86.2} & \cellcolor{gray!15}$+0.2$
      & \textbf{84.2} & \cellcolor{gray!15}$-1.8$
      & \textbf{87.9}
      & \textbf{86.8} & \cellcolor{gray!15}$-1.1$
      & \textbf{88.3} & \cellcolor{gray!15}$+0.4$
      & \textbf{86.6} & \cellcolor{gray!15}$-1.3$
      & \textbf{62.8}
      & \textbf{63.3} & \cellcolor{gray!15}$+0.5$
      & \textbf{62.4} & \cellcolor{gray!15}$-0.4$
      & \textbf{63.3} & \cellcolor{gray!15}$+0.5$ \\

    \bottomrule
  \end{tabular}%
  }
\end{table*}

\paragraph{Implementation details.}
The object encoder is initialized from RI-MAE~\cite{su2025ri}, pretrained on ModelNet40, and fine-tuned end-to-end. Each object is represented by 128 uniformly sampled XYZ points, and RI-MAE-based baselines use the same initialization and sampling. We use a hidden dimension of 256, a 2-layer node-level GNN, and two non-shared 2-layer attention-based relation GNNs for the observer-independent and observer-dependent pathways. All projection MLPs use 4 layers with ReLU activation and dropout of 0.4. The loss weights are set to $\lambda_{\mathrm{obj}}=\lambda_{\mathrm{rel}}=1.0$, $\lambda_{\mathrm{orth}}=0.05$, $\lambda_{\mathrm{aux}}=0.1$, and $\beta=1.0$.

Models are trained for 100 epochs using AdamW with learning rate $1\times10^{-4}$, weight decay $1\times10^{-4}$, and batch size 8, together with cosine decay and a 5-epoch linear warmup. Frame-reorientation augmented baselines use the same optimization settings and are trained until validation performance saturates. The checkpoint with the highest reference-frame validation SGCls mR@50 is selected for each run; reoriented test frames are never used for model selection. Our method does not use frame-reorientation augmentation, and all results are averaged over three random seeds.

\subsection{Main Results}

% We evaluate from three perspectives: standard 3DSGG performance, viewpoint robustness under yaw rotations, and qualitative analysis.
We evaluate from three perspectives: standard 3DSGG performance, relation prediction under observer-frame reorientation, and qualitative analysis.

\paragraph{Standard 3DSGG performance.}
% We first evaluate under the standard unrotated setting to verify that transformation-aware relation learning remains competitive.
We first evaluate under the standard reference-frame setting to verify that observer-aware relation modeling remains competitive with conventional 3DSGG methods.

% As shown in Table~\ref{tab:comparison_results}, our method achieves the best object recognition (60.51 R@1, 81.79 R@5) and competitive predicate and triplet prediction. 
% VL-SAT(RI-MAE) already uses the same RI-MAE encoder as our method; our consistent improvements over it confirm that the gain stems from transformation-aware decoupling rather than the encoder alone. The advantage is further amplified under SGCls and PredCls, where mean recall captures the long-tail predicate distribution.

% As shown in Tables~\ref{tab:comparison_results} and~\ref{tab:sgcls_predcls}, our method remains competitive under the standard unrotated setting. It achieves the best object recognition and the best mean recall in both SGCls and PredCls, while maintaining comparable predicate and triplet recall to existing SOTA methods. 

As shown in Tables~\ref{tab:comparison_results} and~\ref{tab:sgcls_predcls}, our method remains competitive under the standard reference-frame setting. It achieves the best object recognition and mean recall in both SGCls and PredCls, while maintaining comparable predicate and triplet recall to existing state-of-the-art methods. The comparison with VL-SAT+RI-MAE, together with the observer-frame evaluation below, suggests that these gains are not solely attributable to the RI-MAE encoder.

\paragraph{Observer-frame evaluation.}
We next evaluate relation prediction under the cardinal diagnostic subset to assess predicate-specific observer-frame responses.

\begin{figure*}[t]
\centering
\includegraphics[width=\textwidth]{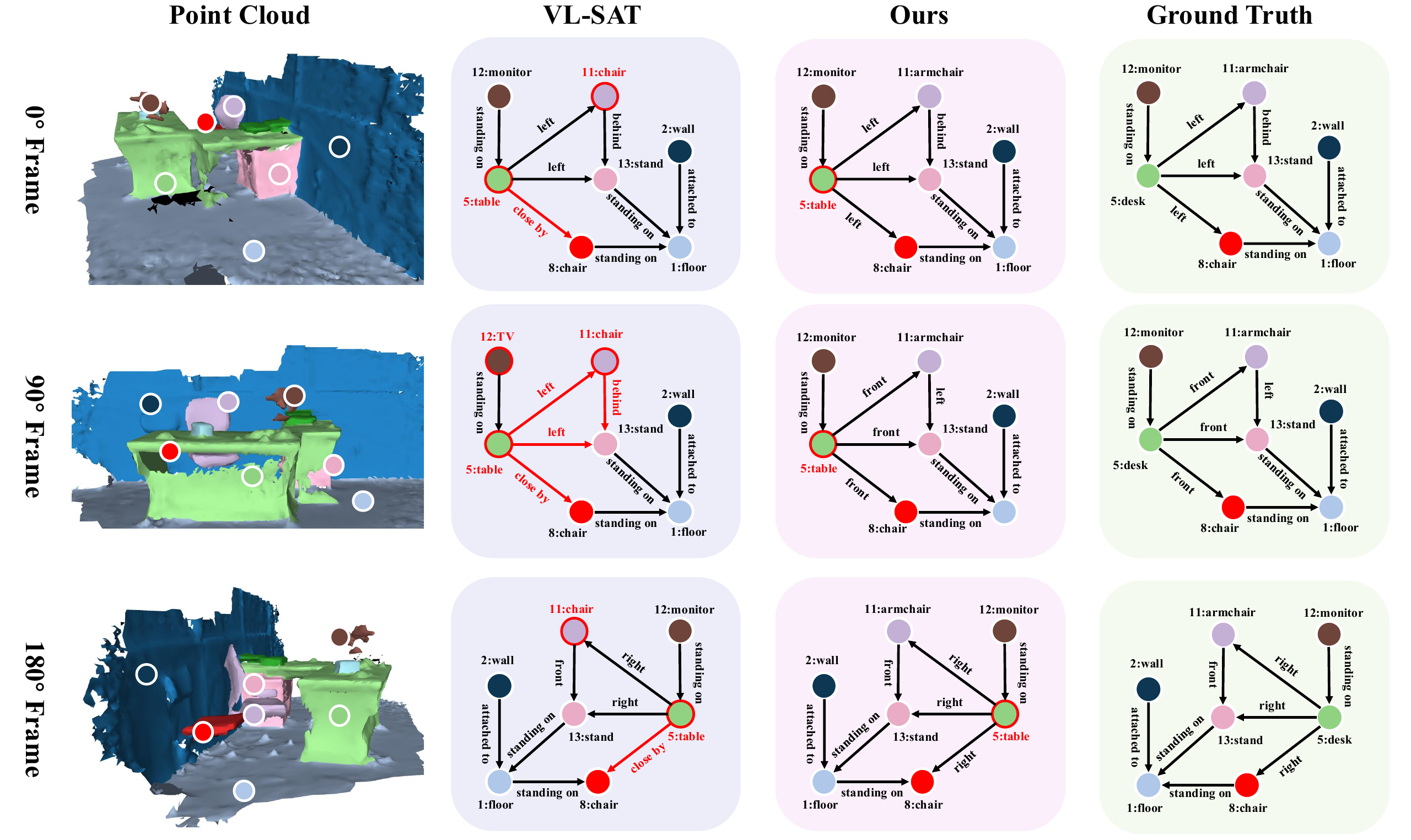}
\caption[Qualitative comparison under observer-frame reorientation.]{Qualitative comparison under observer-frame reorientation at $0^\circ$, $90^\circ$, and $180^\circ$. Red object labels and arrows indicate incorrect object and predicate predictions, respectively. VL-SAT produces plausible predictions in the reference frame but becomes inconsistent under cross-axis reinterpretation. Our method better follows the expected re-indexing of Observer-Dependent Relations (\textit{left}, \textit{right}, and \textit{front}) while preserving Observer-Independent Relations (\textit{standing on} and \textit{attached to}).
}
\label{fig:qualitative_rotation}
\end{figure*}

Table~\ref{tab:rotated_view_robustness} reports relation-focused performance under cardinal observer-frame reorientation. Non-augmented baselines suffer large drops at $90^\circ$ and $270^\circ$; for example, VL-SAT and OCRL decrease by 11.7 and 12.1 Overall R@50 points at $90^\circ$, respectively, while their $180^\circ$ results remain close to the reference frame. This pattern aligns with the $C_4/C_2$ diagnostic distinction: within-axis reversal at $180^\circ$ is comparatively easier to handle, whereas $90^\circ/270^\circ$ additionally require cross-axis reinterpretation. Stronger object representations alone do not resolve this issue, as VL-SAT+RI-MAE still drops by 10.9 points at $90^\circ$. 
Cardinal frame-reorientation augmentation substantially reduces variation across frames, but quadruples the downstream training set and remains clearly below OAR. In particular, augmented baselines achieve lower reference-frame Overall R@50 and lower observer-independent mR@50 than OAR, indicating that augmentation alone does not fully reconcile the distinct frame responses of the two relation categories. In contrast, our method reduces the $0^\circ\!\rightarrow\!90^\circ$ gap to 1.5 points and maintains observer-independent mR@50 between 62.4 and 63.3 across the evaluated cardinal frames, without frame-reorientation augmentation.

% To test whether post-hoc geometric correction alone can resolve this predicate-transformation conflict, we evaluate GeoRule-Hybrid as a counterfactual control. Although GeoRule-Hybrid improves VL-SAT at $90^\circ$ from 68.2 to 74.6 R@50, it reduces canonical-view performance from 79.9 to 76.0 R@50 and still falls substantially below TAD. This result shows that deterministic output correction alone cannot resolve the conflict between yaw-invariant and direction-sensitive predicates; transformation-aware relation representations are necessary. Additional comparisons with prediction-stage group handling, SGCls rotation, and continuous-yaw stress tests are provided in the supplementary material.

To test whether post-hoc geometric correction alone can resolve heterogeneous observer-frame dependency, we evaluate GeoRule-Hybrid as a counterfactual control. Although GeoRule-Hybrid improves VL-SAT at $90^\circ$ from 68.2 to 74.6 R@50, it reduces reference-frame performance from 79.9 to 76.0 R@50 and remains substantially below OAR. This result indicates that deterministic output correction alone cannot reconcile Observer-Independent and Observer-Dependent Relations; observer-aware relation representations are still necessary. Additional comparisons with prediction-stage group handling, SGCls under observer-frame reorientation, and dense-angle evaluations at $30^\circ$ intervals are provided in the supplementary material.

\paragraph{Qualitative results.}
%shows that VL-SAT becomes unstable under 90° axis exchange, while TAD better follows directional transformations and preserves invariant predicates. Additional qualitative results under standard 3DSGG settings are provided in the supplementary material.
Fig.~\ref{fig:qualitative_rotation} shows that VL-SAT becomes unstable under the $90^\circ$ cross-axis reorientation, whereas OAR better follows the re-indexing of Observer-Dependent Relations while preserving Observer-Independent Relations. Additional qualitative results under standard 3DSGG settings are provided in the supplementary material.

% compares representative predictions under $0^\circ$, $90^\circ$, and $180^\circ$ yaw rotations. VL-SAT produces plausible predictions in the canonical view but becomes unstable under $90^\circ$ axis exchange, especially for direction-sensitive predicates such as \textit{left} and \textit{front}. In contrast, our method better follows the expected directional transformation, e.g., \textit{left}$\rightarrow$\textit{front}, while preserving yaw-invariant predicates such as \textit{standing on} and \textit{attached to}. Additional qualitative comparisons are provided in the supplementary material.

\subsection{Ablation Study}
\label{sec:ablation}
% We conduct ablation studies from three perspectives: architectural components, structural decoupling necessity, and objective-level specialization. The baseline uses a PointNet++~\cite{qi2017pointnet++} object encoder and a single GNN relation branch with a standard entangled geometric descriptor. When TAD is applied without TSD, both branches receive the same descriptor but use independent parameters. For rotation robustness, we report $90^\circ$ as the representative axis-exchange condition.
We ablate three components: FSOE, OAGE, and relation specialization. The baseline uses a PointNet++~\cite{qi2017pointnet++} object encoder, a single relation GNN, and a standard unified geometric descriptor. Relation specialization introduces two non-shared relation GNNs, $\mathcal{L}_{\mathrm{orth}}$, and auxiliary supervision. Disabling it restores the single-GNN baseline and removes both supervision terms; without OAGE, the two specialized pathways receive the same descriptor.
% ODS includes two non-shared relation GNNs, $\mathcal{L}_{\mathrm{orth}}$, and auxiliary supervision. When ODS is applied without OAGE, the two pathways receive the same descriptor.

\begin{table}[t]
\centering
\caption{Ablation of components. Spec. denotes non-shared relation GNNs with observer-dependency supervision.}
\label{tab:ablation_components}
\footnotesize
\setlength{\tabcolsep}{3pt}
\renewcommand{\arraystretch}{1.1}
\resizebox{\columnwidth}{!}{%
\begin{tabular}{ccc*{8}{c}}
\toprule
\multirow{2}{*}{FSOE} &
\multirow{2}{*}{Spec.} &
\multirow{2}{*}{OAGE} &
\multicolumn{2}{c}{Triplet} &
\multicolumn{2}{c}{SGCls} &
\multicolumn{2}{c}{PredCls} &
\multicolumn{2}{c}{$90^\circ$ PredCls} \\
\cmidrule(lr){4-5}
\cmidrule(lr){6-7}
\cmidrule(lr){8-9}
\cmidrule(lr){10-11}
& &
& R@50 & mR@50
& R@50 & mR@50
& R@50 & mR@50
& R@50 & mR@50 \\
\midrule

& &
& 89.83 & 60.37
& 31.5 & 23.6
& 79.2 & 57.8
& 59.8 & 46.4 \\

$\checkmark$ & &
& 90.37 & 62.59
& 36.7 & 32.1
& 83.6 & 63.6
& 68.4 & 51.5 \\

$\checkmark$ & $\checkmark$ &
& 90.95 & 67.39
& 37.3 & 33.8
& 85.0 & 64.3
& 83.3 & 62.7 \\

& $\checkmark$ & $\checkmark$
& 89.64 & 62.41
& 33.2 & 28.6
& 80.8 & 62.6
& 78.4 & 61.2 \\

$\checkmark$ & $\checkmark$ & $\checkmark$
& \textbf{91.03} & \textbf{69.85}
& \textbf{38.1} & \textbf{34.6}
& \textbf{86.0} & \textbf{66.7}
& \textbf{84.5} & \textbf{66.9} \\

\bottomrule
\end{tabular}%
}
\vspace{-3mm}
\end{table}

Table~\ref{tab:ablation_components} shows the contribution of each component. FSOE improves standard PredCls from 79.2/57.8 to 83.6/63.6 in R@50/mR@50 and raises the $90^\circ$ result from 59.8/46.4 to 68.4/51.5, showing that rotation-robust object features provide a stronger basis but remain insufficient for cross-axis reinterpretation. Adding relation specialization produces the largest gain, reaching 83.3 R@50 and 62.7 mR@50 at $90^\circ$. 
Combining relation specialization and OAGE without FSOE still yields 78.4/61.2, showing that the relation-level design remains effective with a standard object encoder. The full model achieves the best results across standard and reoriented settings, reaching 84.5 R@50 and 66.9 mR@50 at $90^\circ$. Further ablations show that sharing pathway parameters or removing $\mathcal{L}_{\mathrm{orth}}$ or $\mathcal{L}_{\mathrm{aux}}$ degrades performance, complementing the OAGE ablation above.
\section{Conclusion}

% We presented TAD, a transformation-aware framework for viewpoint-robust 3D scene graph generation under yaw changes. Our key insight is that yaw rotations impose conflicting requirements on yaw-invariant and direction-sensitive predicates that a single entangled relation representation cannot resolve. By decoupling relation reasoning through transformation-specific descriptors, parameter-decoupled branches, and group-aware auxiliary supervision, TAD achieves state-of-the-art mean recall on standard SGCls/PredCls and robust rotated-view predicate prediction without training-time rotation augmentation.
% We presented OAR, an observer-aware framework for 3D scene graph generation under observer-frame reorientation. Our key insight is that Observer-Independent and Observer-Dependent Relations require distinct responses to changes in the observer frame, which are difficult to capture with a single frame-agnostic relation representation. 
% By aligning relation reasoning with these heterogeneous dependencies through observer-aware geometric encoding, relation-specialized context modeling, and observer-dependency supervision, OAR achieves state-of-the-art mean recall on standard SGCls and PredCls, while maintaining reliable predicate prediction across reoriented observer frames without frame-reorientation augmentation.

We presented OAR, an observer-aware framework for 3D scene graph generation under observer-frame reorientation. Our key insight is that Observer-Independent and Observer-Dependent Relations require distinct responses to changes in the observer frame. OAR models these heterogeneous dependencies through observer-aware geometric encoding and relation specialization, supported by frame-stable object encoding, while preserving unified multi-label predicate prediction. It achieves the best mean recall among the compared methods on standard SGCls and PredCls and maintains reliable relation prediction across reoriented observer frames without frame-reorientation augmentation.
% OAR combines frame-stable object encoding, observer-aware geometric encoding, and relation specialization to learn complementary observer-independent and observer-dependent cues for unified multi-label predicate prediction. It achieves the best mean recall among the compared methods on standard SGCls and PredCls, while maintaining reliable predicate prediction across reoriented observer frames without frame-reorientation augmentation.
% Our evaluation establishes the effectiveness of TAD for yaw-frame robustness on the closed-vocabulary 3DSSG benchmark with standard instance masks. Extending this controlled setting to partial observations, occlusions, and reconstruction noise is an important next step toward end-to-end embodied deployment. Moreover, TAD currently uses a manually specified directional/invariant predicate partition; learning transformation groups from data may further broaden its applicability to richer relation vocabularies.
% Our study focuses on yaw-frame robustness in closed-vocabulary 3DSSG. Extending TAD to broader sensing conditions and learned predicate transformation groups is left for future work.
% Our study focuses on yaw-frame robustness in closed-vocabulary 3DSSG. Broader sensing conditions and learned predicate transformation groups remain future work.

Our study focuses on controlled observer-frame reorientation in closed-vocabulary 3DSSG. Broader sensing conditions and data-driven modeling of predicate-specific observer-frame dependency remain future work.

\bibliography{main}

\appendix

{
    \small
    \bibliographystyle{ieeenat_fullname}
    %\bibliography{main} % Assuming your .bib file is named main.bib
}

% WARNING: do not forget to delete the supplementary pages from your submission for the main paper
%\input{sections/supplementary}

\end{document}